\pdfoutput=1

\documentclass[11pt]{article}

\usepackage[]{EMNLP2023}

\usepackage{times}
\usepackage{latexsym}
\usepackage{amssymb}
\usepackage{amsmath}
\usepackage{graphicx}
\graphicspath{{images/}}

\usepackage{amsthm,amsmath,amssymb}
\usepackage{bbm}

\usepackage{mathrsfs}
\usepackage{booktabs}
\usepackage{multirow}
\usepackage{makecell}

\usepackage{xcolor}

\usepackage[T1]{fontenc}

\usepackage[utf8]{inputenc}

\usepackage{microtype}

\usepackage{inconsolata}

\newcommand\blfootnote[1]{%
  \begingroup
  \renewcommand\thefootnote{}\footnote{#1}%
  \addtocounter{footnote}{-1}%
  \endgroup
}
\usepackage[linesnumbered,ruled,lined,noline,noend]{algorithm2e}
\SetKwIF{If}{ElseIf}{Else}{if}{}{else if}{else}{end if}%
\SetKwFor{For}{for}{}{end for}%

\title{Synslator: An Interactive Machine Translation Tool \\ with Online Learning}

\author{Jiayi Wang$^\ast$\textsuperscript{1}, Ke Wang$^\ast$\textsuperscript{2}, Fengming Zhou\textsuperscript{2}, Chengyu Wang\textsuperscript{3}, \\ \textbf{Zhiyong Fu\textsuperscript{2},} \textbf{Zeyu Feng\textsuperscript{2},} \textbf{Yu Zhao\textsuperscript{2},} \textbf{Yuqi Zhang\textsuperscript{2}} \\
\textsuperscript{1}University College London, \textsuperscript{2}Alibaba DAMO Academy, \textsuperscript{3}Alibaba Cloud\\
\texttt{ucabj45@ucl.ac.uk,\{wk258730,zfm104435,chengyu.wcy\}@alibaba-inc.com,}\\
\texttt{\{zhiyong.fzy,zeyu.fz,kongyu,chenwei.zyq\}@alibaba-inc.com}
}

\begin{document}
\maketitle
\blfootnote{$^\ast$ indicates equal contributions.}
\begin{abstract}
Interactive machine translation~(IMT) has emerged as a progression of the computer-aided translation paradigm, where the machine translation system and the human translator collaborate to produce high-quality translations. 
This paper introduces Synslator, a user-friendly computer-aided translation~(CAT) tool that not only supports IMT, but is adept at online learning with real-time translation memories. 
To accommodate various deployment environments for CAT services, Synslator integrates two different neural translation models to handle translation memories for online learning. Additionally, the system employs a language model to enhance the fluency of translations in an interactive mode. 
In evaluation, we have confirmed the effectiveness of online learning through the translation models, and have observed a 13\% increase in post-editing efficiency with the interactive functionalities of Synslator. A tutorial video is available at:~\url{https://youtu.be/K0vRsb2lTt8}.
\end{abstract}

\section{Introduction and Related Works}
We have witnessed consistent advancements made in the field of machine translation~\citep{lopez2008statistical,koehn2009statistical,sutskever2014sequence,bahdanau2014neural,vaswani2017attention}, which progressively enhance the quality of translations. These continuous advancements have prompted a transformation in the translation industry, with a shift from exclusive reliance on human translators to the integration of computer-aided translation (CAT) methods~\citep{bowker2002computer,bowker2010computer,green2013efficacy,laubli2013assessing,bowker2014computer}. For CAT, instead of translating from scratch, human translators engage in post-editing tasks, refining machine translation outcomes to yield the final approved results, and thus considerably improving the translation quality. 

The post-editing process used to be generally static, wherein machines would cease to respond to human modifications as soon as human post-editing began~\citep{bowker2010computer,green2013efficacy}.
Recent studies have explored interactive procedures and algorithms~\citep{green2015natural,knowles-koehn-2016-neural,santy2019inmt,chatterjee2019automatic,wang2020computer,wang2020alibaba,ge2022tsmind,wang2022easy}, enabling a more collaborative process between humans and machines, where machines can dynamically adjust translations in line with the edits made by humans. 

Translation Memory~(TM) is a key component that can be optimally leveraged within the realm of CAT~\citep{green2014predictive}. As human translators undertake post-editing with CAT tools, incremental online TMs can be invariably accumulated. Hence, the capability to use TMs for online learning emerges as a critical attribute for CAT~\citep{wang2022non}. In fact, there are different environment settings for the deployment of CAT services. In environments where the deployment of CAT allows for authorized usage of TMs, it is feasible to utilize translation memories for model training~\citep{bulte2019neural,xu2020boosting,bapna2019non}. While in different settings such as public cloud solutions for CAT services, translation memories are usually introduced online by users, and we are not authorized to train models using them. Then, it becomes more beneficial to have a translation model capable of handling online TMs during the inference phase. For instance, \citet{khandelwal2020nearest} predicts target words with a $k$-nearest-neighbor ($k$NN) classifier over a datastore of cached TM examples during inference.

\begin{figure*}[t]
\centering
\includegraphics[width=0.9\textwidth]{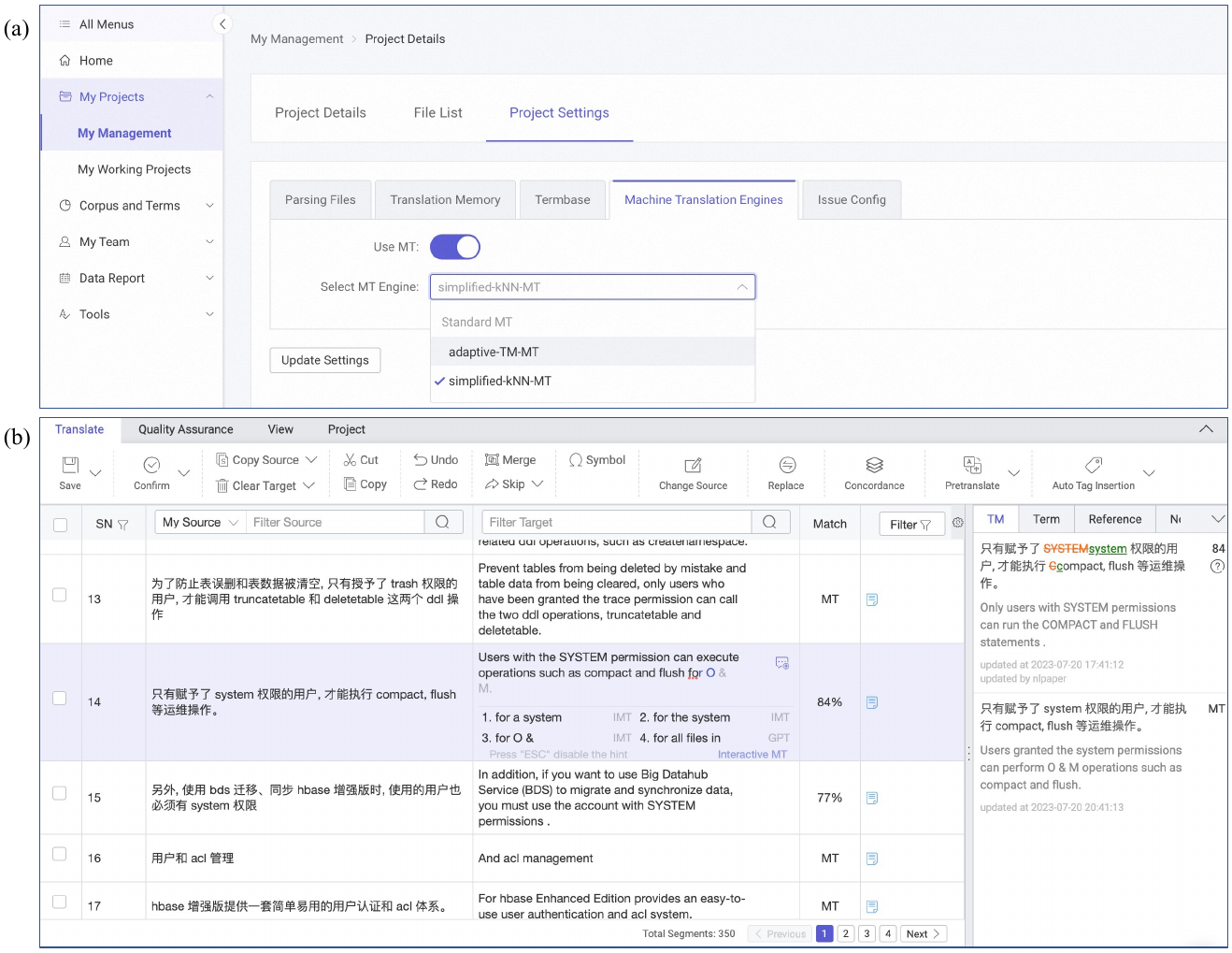}
\caption{\small{The user interfaces of Synslator: (a) the project setting interface, (b) the post-editing interface.}}
\label{fig:combine_ui}
\end{figure*}

In this paper, we present Synslator, a tool designed for computer-aided translation that supports interactive machine translation through the application of a subword-prefix decoding algorithm. 
This tool allows human translators to receive automated translation suggestions in real-time while editing machine translations as needed and grants users the flexibility to make edits at the character/subword level. To cater diverse deployment requirements of CAT services as aforementioned, Synslator employs two distinct models to handle translation memory for online learning. These include an adaptive neural machine translation model named~\textit{adaptive}-TM-MT, and a nearest-neighbor-retrieval based machine translation model, which we call~\textit{simplified}-$k$NN-MT. Moreover, Synslator additionally produces suggestions that are purely grounded by a GPT-based language model (LM) from the perspective of monolingual fluency and styling, which serve as supplementary references for human translators. 
The subword-prefix decoding algorithm can accommodate all of these models given human's subword-prefix inputs.

\section{Synslator: The Proposed System}
In this section, we will introduce the major functionalities of Synslator and the algorithm implementations behind this tool. 
The fundamental feature of Synslator is to allow users to create a translation project, configure its respective settings, and perform post-editing based on machine translation results~\footnote{A tutorial video, available at this link:~\url{https://youtu.be/K0vRsb2lTt8}, demonstrates how to create a new project. The example project shown in the video focuses on legal translation from Chinese to English. We presume in this case that the CAT tool is set up as a public cloud service, and the user has already uploaded a suitable TM dataset.}.
As depicted in the screenshots in Figure~\ref{fig:combine_ui}, there are two user interfaces that human translators utilize to finish a translation project. The interface (a) allows adjustments for project settings, while the interface (b) supports human post-editing.

\begin{figure}[t]
\centering
\includegraphics[width=0.48\textwidth]{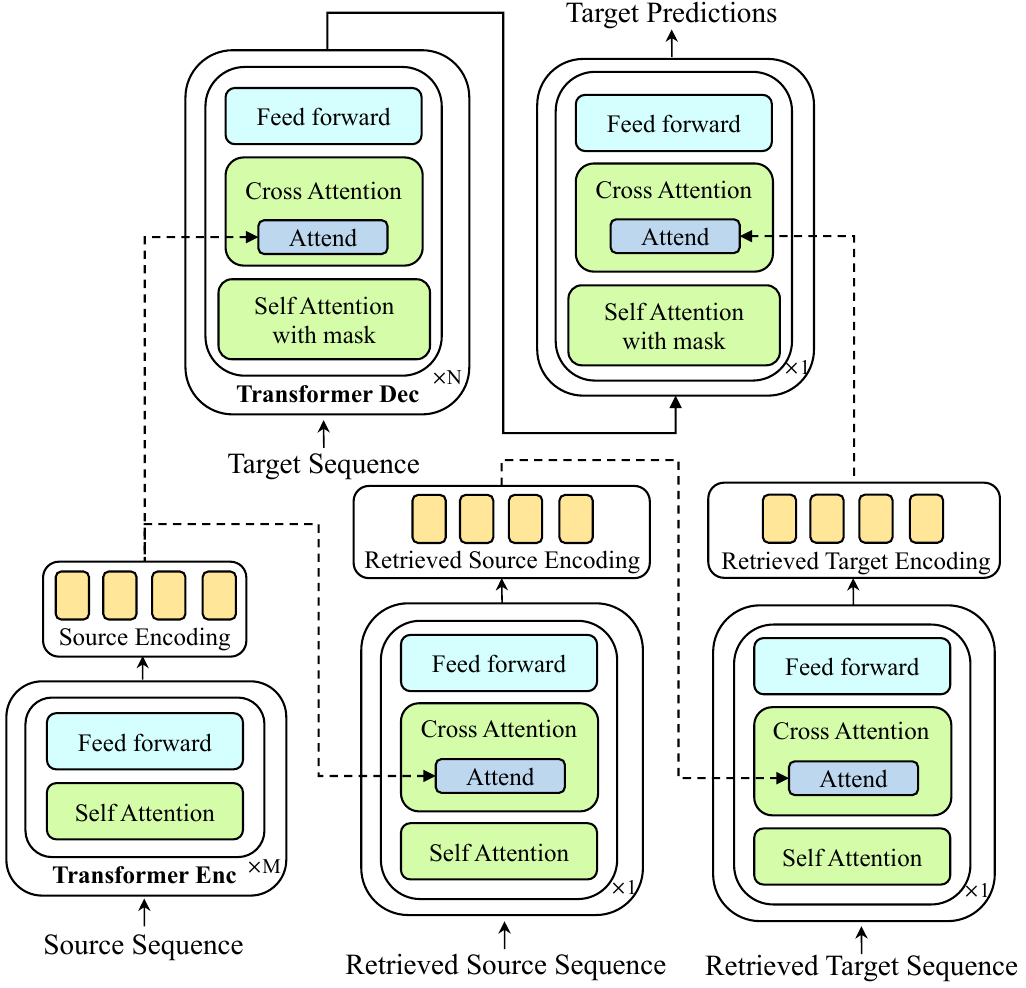}
\caption{\small{The~\textit{adaptive}-TM-MT framework, featuring dashed lines to represent cross-attention.}}
\label{fig:TM-MT}
\end{figure}

\subsection{Project Setting Interface}
Users are presented with choices for file parsing, selection of translation memory, selection of termbase, and the option to choose different machine translation engines. The file parsing functionality is employed to segment sentences if a document is uploaded for translation. We will focus on functions of translation memory, termbase and machine translation engines.

\subsubsection{Translation Memory and Termbase} 
\label{sec:TM_retrieve}
After creating a translation project, users can upload related TMs and bilingual termbase. For each source sentence to be translated, Synslator will present the most relevant TM including its source and target translation for human reference in the post-editing interface. The process of searching for similarity is initially carried out by an open-sourced distributed search engine, ElasticSearch~\footnote{https://github.com/elastic/elasticsearch}, which retrieves as most as 64 bilingual sentence pairs from the source side that exhibit the highest relevance scores. Subsequently, from these bilingual sentence pairs, we select the one demonstrating the most similarity based on the computation of the edit distance on the source side as well. The minimum threshold for the edit distance is denoted as the Minimum Match Rate, and its value can be set in the Project Settings interface. 
When it comes to translating terms present in the source sentence, Synslator simply utilizes an exact match strategy to locate their respective translations from the bilingual termbase. If multiple matches are found, all of them will be displayed in the post-editing interface.

\begin{figure*}[t]
\centering
\includegraphics[width=0.7\textwidth]{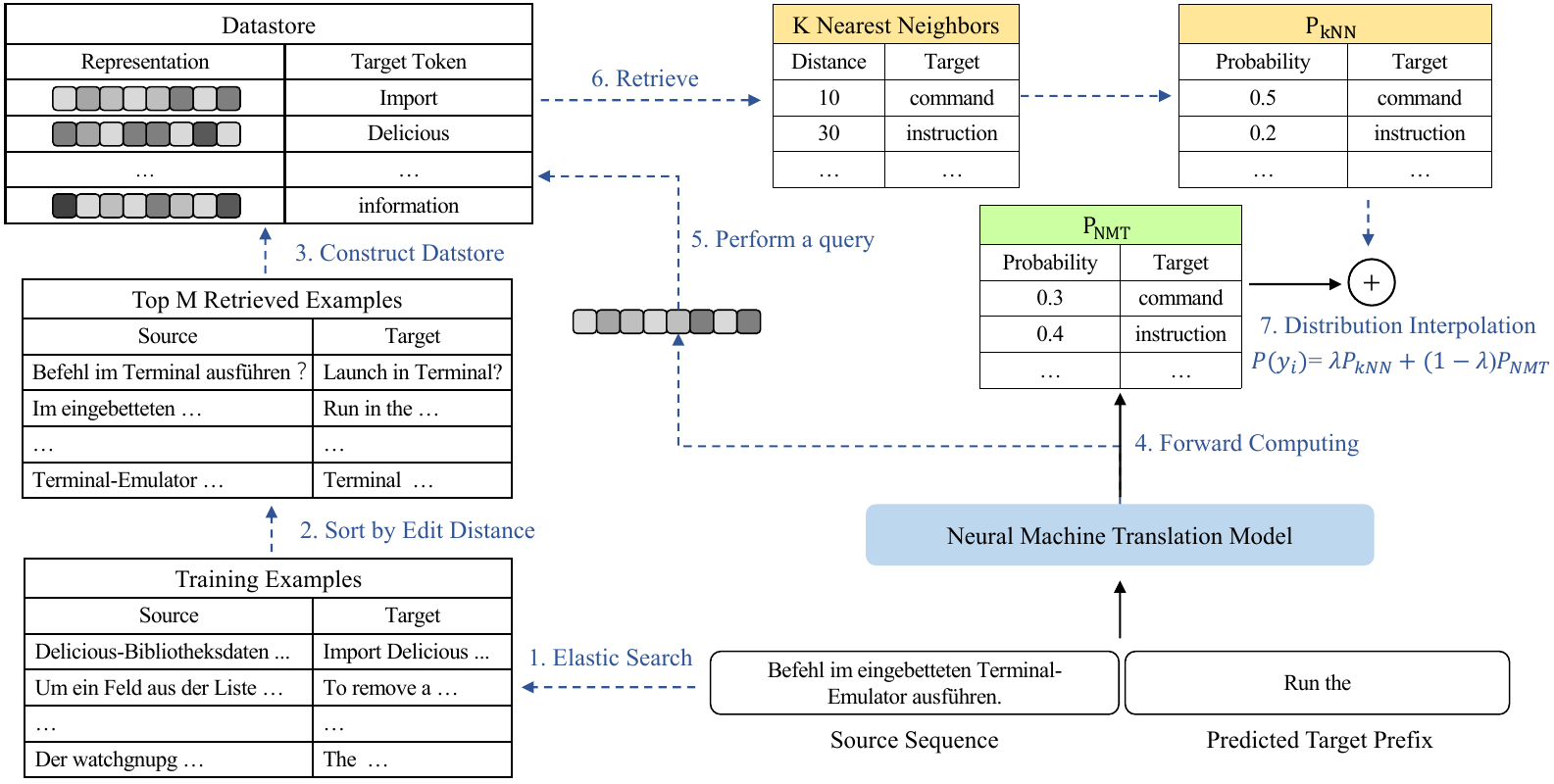}
\caption{\small{The~\textit{simplified}-$k$NN-MT framework with sequentially numbered workflow steps.}}
\label{fig:knn-mt}
\end{figure*}

\subsubsection{Machine Translation Engines}
\label{sec:engine}
As discussed, the process of post-editing with the CAT tool results in incremental online TMs. In response to the varying deployment environments associated with CAT services, Synslator utilizes two distinct models for online learning accordingly.

\paragraph{\textit{adaptive}-TM-MT}
When the usage of training TMs is granted, such as in establishing CAT services for private deployment, it becomes particularly advantageous to utilize TMs to boost the model performance in domain-specific translation via fine-tuning. For instance, beyond using parallel bilingual training data, it is feasible to employ related TMs as an additional input to the model for further enhancement. \citet{bapna2019non} retrieves neighbors from TMs and incorporates them into the model through Conditional Source Target Memory. Inspired by their work, we propose the~\textit{adaptive}-TM-MT as illustrated in Figure~\ref{fig:TM-MT}. 

Given a pre-trained Transformer model, we fine-tune it with TMs as domain-specific training data. For each parallel sentence pair of TMs, we first use the pre-trained encoder and decoder to encode the TM's source and target sequences; afterwards, we execute a retrieval process to locate the nearest neighbor from the remaining TMs in the same way as in Section~\ref{sec:TM_retrieve}. The retrieved source-target pair is subsequently encoded with additional Transformer layers. More specifically, the retrieved source is encoded within one Transformer encoder layer, and we integrate information from the encoder representation of the source sequence by using a cross-attention with it. The retrieved target is then encoded in a similar fashion, attending to the encoded retrieved source memory. Instead of using Gated Multi-Source Attention from~\citet{bapna2019non}, we simply add one Transformer decoder layer upon the original decoder module, attending the encoded retrieved target memory. Once the~\textit{adaptive}-TM-MT is trained offline with historical TMs, it would obtain the capability of handling incremental TMs for online learning in CAT. 


\paragraph{\textit{simplified}-$k$NN-MT}
When training with TMs is not authorized, such as for public cloud solutions for CAT services where translation memories are commonly loaded online, it would be more effective to develop a model capable of handling plug-in TMs during inference. Motivated by $k$NN-MT~\citep{khandelwal2020nearest}, we propose the~\textit{simplified}-$k$NN-MT, a pared-down variant of $k$NN-MT, with its architecture displayed in Figure~\ref{fig:knn-mt}. 

In comparison with $k$NN-MT, we simplify the process of datastore construction with all TMs, and instead, we adopt the TM matching approach detailed in Section~\ref{sec:TM_retrieve} to obtain a smaller amount of TMs for datastore construction. For each source sentence, we collect up to 16 of the most pertinent neighboring TMs after ElasticSearch, identified through edit distance, to create a condensed datastore for $k$NN search. This optimized procedure effectively alleviates computational complexity and storage requirements, thereby enhancing its applicability in practical scenarios. It is important to note that every source sentence due for translation is associated with a unique datastore. Once the datastore is constructed, the model can predict target words by interpolating the distribution of $k$NN predictions in the same manner as $k$NN-MT. 

However, the condensed datastore could introduce noise when retrieving $k$NN based on representation distance for each target prediction. To mitigate the potential impact of such noise on translation performance, we set a maximum distance threshold, represented by $\tau$. Only those neighbors with a distance smaller than $\tau$ will be selected. If none meets this condition, there will be no $k$NN interpolation for generation. In scenarios of public cloud solutions for CAT services, all hyper-parameters including the newly introduced $\tau$ can be tuned on the domain-specific development sets.

\subsection{Post-Editing Interface}
Upon proper configuration of the project settings, we can enter the workbench to access the post-editing interface, which is depicted in the screenshot (b) in Figure~\ref{fig:combine_ui}. This interface displays all static machine translation results before modifications by human translators. 
As soon as human post-editing begins, the translation model will provide refined translation results based on human edits. The human translator can continually make adjustments, while the translation model also consistently refines its outputs based on human inputs. This iterative process of interaction endures until the translation meets the quality standard. 

\subsubsection{Workflow of Post-Editing}
\paragraph{Online Learning with TM} 
The corresponding translation memory and termbases, as described in Section~\ref{sec:TM_retrieve}, are displayed on the right side of the post-editing interface. Human translators can double-click these resources for immediate inclusion in the translation result, and make any modifications to them as required. More importantly, incremental online TMs gathered through human post-editing will be merged in real-time with previous memories, which establishes favorable conditions for the~\textit{adaptive}-TM-MT and~\textit{simplified}-$k$NN-MT models to facilitate online learning. Regarding the~\textit{adaptive}-TM-MT, the source sentence and the highest-ranking matching TM from the current total memories are fed into the model for generations. In the case of the~\textit{simplified}-$k$NN-MT, for every source sentence, Synslator will gather at most 16 relevant TMs to construct a condensed datastore for $k$NN retrievals during $k$NN-MT inference. 

\paragraph{Translation Refinement} Human adjustments can be made at a character/subword level. The translation model, guided by the subword input from human translators and all previously generated target words, automatically completes the current target word and generates adjusted subsequent words to finalize a translation. For example, as shown in (b) of Figure~\ref{fig:combine_ui}, given the source sentence with id 14 and its original machine translation result, the human translator deletes the original words following ``flush'' and enters two characters ``fo''. Immediately, the translation model automatically completes the word ``for'' and generates subsequent target words. This mechanism is facilitated by our proposed subword-prefix decoding algorithm, which will be introduced in Section~\ref{sec:subword_prefix_dec}. 

\paragraph{Prediction Highlighting with Likelihood}When human translators verify the correctness of the translation up to a certain target word, they can click on the spot to lock in the preceding translated text. Besides, word predictions will be sequentially highlighted with a separate color as long as their translation probability remains high (e.g., above 0.6), indicating the confidence of the model. 
In the same example in (b) of Figure~\ref{fig:combine_ui}, the ``O'' following ``for'' is highlighted, which the translation model believes to be a highly likely correct word prediction. If human translators agree with the correctness of the highlighted predictions, they can use the TAB key to swiftly secure them. 

\paragraph{Suggestion Box} Beneath each translated sentence, a suggestion box is featured. It furnishes the next 3-best translations generated by the translation model, excluding the highest-ranked one which is already displayed.
The box also includes a suggestion derived from a GPT-based LM. This acts as a supplementary reference, offering insights into monolingual fluency and stylistic nuances. However, the precision of a GPT model's next-word prediction depends on the preceding context~\citep{lagler2013gpt2,floridi2020gpt}. The LM only provides a suggestion when the target prefix composes of more than ten translated words. In our design, 3-best translation suggestions are limited to three-word predictions, while the LM predicts the next four words. The 3-best translation suggestions are de-duplicated for clarity.

\subsubsection{Subword-Prefix Decoding}
\label{sec:subword_prefix_dec}
\begin{figure}[t]
\centering
\includegraphics[width=0.45\textwidth]{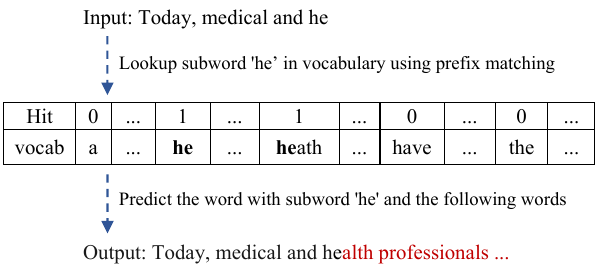}
\caption{\small{An example of Hit Vector.}}
\label{fig:subword}
\end{figure}

During post-editing, when the last input from the human translator is a space, it indicates the presence of a fully-formed word preceding the space character. In this case, both the translation model and the GPT-based LM can anticipate the ensuing words through the application of a forced decoding mode. 
Otherwise, given the subword prefix from human inputs, we build a binary vector with the target vocabulary size, called Hit Vector, by looking up the subword prefix in the vocabulary using exact matching. In this vector, any index with a value of 1 represents a match with the subword prefix, denoting a successful ``hit'' by the subword prefix. An example of Hit Vector is illustrated in Figure~\ref{fig:subword}. Among the words that the subword prefix hits, our model selects the one with the highest generation probability as the current word prediction. Subsequently, our model autoregressively generates all the following words using a forced decoding mode. The subword-prefix decoding algorithm not only facilitates the beam search decoding in translation models, 
but also supports the ``top-$k$ sampling'' decoding strategy prevalent in the GPT-based LM~\footnote{As an example, the pseudo codes of the subword-prefix decoding algorithm for translation models are displayed in Algorithm~\ref{alg:beam_search} of Appendix~\ref{sec:appendix}.}.
Please note that due to varied lengths of target prefix sequences, we employ a strategy of left-padding for batch decoding to enable parallel computation.

\section{Evaluation}
We implement assessment protocols to measure the performance of Synslator from two perspectives: (1) evaluating the effectiveness of online learning for domain-specific translation tasks;
and (2) assessing real-time post-editing efficiency with interactive functionalities. 

\subsection{Evaluation for Online Learning}
\paragraph{Datasets and Model Settings}
We utilize an in-house pre-trained Chinese-English neural machine translation~(NMT) model to facilitate the implementation of the~\textit{adaptive}-TM-MT and the~\textit{simplified}-$k$NN-MT. This NMT model is based on the Transformer architecture~\cite{10.5555/3295222.3295349}.
For the GPT-based LM, we follow the same architecture as the decoder of the pre-trained NMT model and train it with in-house English monolingual data from scratch. 

We employ an in-house Chinese-English IT-domain training set to train the~\textit{adaptive}-TM-MT. Its encoder and decoder are initialized with the pre-trained NMT model, while other layers are trained from scratch. Once the training reaches convergence, we proceed to evaluate online learning using the corresponding test set, where the training data serve as TMs for retrievals.
In evaluating the~\textit{simplified}-$k$NN-MT, we use open-sourced Chinese-English Law and Subtitle in-domain training sets~\cite{tian-etal-2014-um} as TMs for building condensed datastores and conducting $k$NN retrievals. We equip the pre-trained NMT model with the $k$NN search functionality during inference on the test sets, with all hyper-parameters tuned on the development set, as shown in Table~\ref{tab:knn_param} of Appendix~\ref{sec:appendix}. The statistics of the in-domain datasets are presented in Table~\ref{tab:statistics} of Appendix~\ref{sec:appendix}.

\paragraph{Evaluation Metrics}
The translation outcomes are assessed through both static and dynamic methods. Static translation results are evaluated over the BLEU score using ``multi-bleu.perl'' script of the Moses\footnote{http://statmt.org/moses}. Moreover, we introduce a metric called N-gram Accuracy to assess prediction accuracy given dynamic target prefix inputs. In detail, we enumerate target prefix sequences from the golden references which are assumed as inputs from human translators and enable the translation model to produce predictions for the subsequent N words. N-gram Accuracy is computed by determining the proportion of correct N-gram predictions relative to the total count of N-gram references, i.e.,
\begin{equation}
\small
Acc_{N-gram}=\frac{Count(Pred_{N-gram}=Ref_{N-gram})}{Count(Ref_{N-gram})},
\end{equation}
where $Pred_{N-gram}$ and $Ref_{N-gram}$ are the N-gram prediction and the N-gram reference given the target prefix input. A higher value of $Acc_{N-gram}$ signifies better performance.
\paragraph{Experimental Results}
The experimental results are presented in Table~\ref{tab:results}. We compare the~\textit{adaptive}-TM-MT with the pre-trained NMT model, indicated as Vanilla NMT, and the model with classic fine-tuning~\cite{freitag2016fast}, labeled as FT-NMT. It is clear that the~\textit{adaptive}-TM-MT demonstrates superior performance. It is also evident that the provision of the next 3-best translation suggestions, along with the LM suggestion, further boosts the N-gram Accuracy, demonstrating the importance of the translation box. On the other hand, it is noteworthy that the~\textit{simplified}-$k$NN-MT significantly outperforms the pre-trained NMT model in both of the Law and Subtitle domains. In short, both the~\textit{adaptive}-TM-MT and the~\textit{simplified}-$k$NN-MT can enable online learning effectively.

\subsection{Evaluation of Interactive Functionalities}
We conduct real-time post-editing experiments to assess the efficiency of the interactive functionalities, which mainly include the sub-word prefix decoding and the Suggestion Box. Ten independent translators, each with an 8-year experience in translating Chinese-English in IT-related fields, are randomly split into two groups. Both groups participate in post-editing on identical IT-domain translation projects over a span of three weeks, utilizing the~\textit{adaptive}-TM-MT model for online learning. Projects are assigned randomly among the translators in either group. For comparison, one group (referred to as the MT-PE group) performs static post-editing only with matching TMs and termbase displayed, while the other group undertakes post-editing with Synslator. The statistics of the projects and the total elapsed time of human labor are summarized in Table~\ref{tab:efficiency}. It has been observed that for a total of 213,976 words translated, the efficiency of post-editing is improved by 13\% with the interactive functionalities of Synslator.

\begin{table}[t]
  \centering
\resizebox{\hsize}{!}{
  \begin{tabular}{ l | c | c c c }
    \toprule[1pt]	

   {Model} & {BLEU} &  {$Acc_{1-gram}$} &  {$Acc_{2-gram}$} &  {$Acc_{3-gram}$}  \\
 \midrule[0.75pt]
  \multicolumn{1}{c}{}   &  \multicolumn{4}{c}{In-house IT domain test set} \\
 \midrule[0.75pt]
   {Vanilla NMT} & 25.47 &46.38 &30.47 &21.13  \\ 
   {FT-NMT } & 28.28 &51.63 &35.64 &25.59 \\
 \midrule[0.75pt]
   {\textit{adaptive}-TM-MT} & \textbf{29.05} &52.25 &36.31 &26.30\\
   {~+Next 3-best} & - & 57.18 & 42.13 & 31.87 \\
   {~+Next 3-best and LM } & - & \textbf{62.60}& \textbf{44.07} & \textbf{32.45} \\
 \midrule[0.75pt]
 \midrule[0.75pt]
 \multicolumn{1}{c}{}   &  \multicolumn{4}{c}{Open-sourced Law domain test set} \\
  \midrule[0.75pt]

   {Vanilla NMT} & 33.73 &53.66 &38.11 &28.06 \\
   {\textit{simplified}-$k$NN-MT} & \textbf{37.21} & \textbf{56.45} &\textbf{42.43} & \textbf{33.17} \\
 \midrule[0.75pt]
  \multicolumn{1}{c}{}  &  \multicolumn{4}{c}{Open-sourced Subtitle domain test set} \\
   \midrule[0.75pt]

   {Vanilla NMT} & 18.81 &38.90 &21.95 &12.97 \\
   {\textit{simplified}-$k$NN-MT} & \textbf{20.04} &\textbf{40.61} &\textbf{24.30} &\textbf{15.34} \\
   \bottomrule[1pt]

  \end{tabular}
}

 \caption{\small{Evaluation of translation results using BLEU Scores and N-gram Accuracy.}
 }
  \label{tab:results}
\end{table}

\begin{table}[t]
  \centering
  \small
  \renewcommand\arraystretch{1.2}
\resizebox{\hsize}{!}{
  \begin{tabular}{ c c | c c | c c | c}
    \toprule[1pt]
    \multicolumn{2}{c}{} & \multicolumn{2}{c}{MT-PE} &\multicolumn{2}{c}{Synslator} \\
   \midrule[0.75pt]
   {Week} & {\#Word} & {Time (h)} & {Avg.}  & {Time (h)} & {Avg.} & {Impr.}\\
   \midrule[0.75pt]
   {1} & 90,344 & 147.85 & 611.05 & 136.67 & 661.04 & +8\% \\ 
   {2} & 78,882 & 148.51 & 531.16 & 124.97 & 631.21 & +19\% \\
   {3} & 44,750 & 80.05 & 559.03 & 71.36 & 627.10 & +12\% \\
    \midrule[0.75pt]
   {Total} & 213,976 & 376.41 & 568.46 & 333.00 & 642.57 & +13\% \\
   \bottomrule[1pt]
   \bottomrule[1pt]
  \end{tabular}
}

 \caption{\small{The efficiency of real-time post-editing. The symbols \#Word and Avg. represent the total number of source words in the translation projects and the average number of words completed per hour, respectively.}}
  \label{tab:efficiency}
\end{table}

\section{Conclusion}
We have presented Synslator, a user-friendly IMT tool. In different deployment environments, it utilizes distinct translation models for online learning with real-time translation memories, and provides multiple translation suggestions through a subword-prefix decoding algorithm. In practical applications, Synslator assists human translators to perform efficient post-editing interactively, enhancing the overall translation workflow. 
\bibliography{anthology,custom}
\bibliographystyle{acl_natbib}

\appendix
\section{Appendix}
\label{sec:appendix}
In the appendix section, sample pseudo codes for the subword-prefix decoding algorithm in translation models are showcased in Algorithm~\ref{alg:beam_search}, which has been detailed in Section~\ref{sec:subword_prefix_dec}. Additionally, the paper's supplementary materials include the statistics of the in-domain datasets utilized for evaluating the translation models employed by Synslator, and the hyper-parameter configurations for the~\textit{simplified}-$k$NN-MT, which can be found in Table~\ref{tab:knn_param} and Table~\ref{tab:statistics} respectively.

\begin{algorithm}[ht]
\small
\caption{\small{Subword-Prefix Decoding}}
\label{alg:beam_search}
\SetKwInOut{KwIn}{Input}
\KwIn{$x^{raw}$: source sequence \\
      $y^{raw}$: target sequence with subword-prefix \\
      $tok(.)$: tokenization \\
      $bpe(.)$: Byte-Pair Encoding \\
      $v$: joint vocabulary \\
      }
$x^{tok} \gets tok(x^{raw}), y^{tok} \gets tok(y^{raw})$\;
$x^{bpe} \gets bpe(x^{tok}), y^{bpe} \gets bpe(y^{tok})$\;
\If{$y^{raw}$.lastchar() is not space char $:$}
{
    $hit \gets [0,0,...,0]$\;
    \For{$i \gets  1,2,\ldots,|v|:$}
    {
        \If {$v[i].startwith(y^{bpe}.last()):$}
        {
            $hit[i] \gets 1$\;
        }
    }
    $prob_{|y^{bpe}|} \gets p_{mt}(x^{bpe},y^{bpe})$\;
    $next~token \gets (prob_{|y^{bpe}|} \circ hit).top(1)$\;
}
Proceed with the standard beam search procedure.
\end{algorithm}

\begin{table}[ht!]
  \normalsize
  \centering
  \small
{
\begin{tabular}{c|c c c c}
  \toprule[1pt]	
     {Domain} &  {$k$} & {$\lambda$} & {$Temperature$} & ${\tau}$ \\
  \midrule[0.75pt]
    {Law \& Subtitle } & 4 & 0.4 & 5 & 5 \\
  \bottomrule[1pt]
\end{tabular}

}
  \caption{The hyper-parameters of the~\textit{simplified}-$k$NN-MT, which are tuned on the development sets.}
  \label{tab:knn_param}
\end{table}

\begin{table}[ht!]
  \normalsize 
  \centering
  \small
{
  \begin{tabular}{ l | c |c c }
  \toprule[1pt]	

  {Domain} & {In-house IT}& {Law}  &{Subtitle}
  \\
  \midrule[0.75pt]
     {Train} & {2,300,000}  & {218,000}  & {298,000}
  \\
     {Development} & {2,000} & {2,000} & {2,000}
  \\
     {Test} &{2,000} & {456} & {597}
  \\
  \bottomrule[1pt]
  \end{tabular}
 }
  \caption{The number of parallel sentences in the in-domain datasets employed in assessing the~\textit{adaptive}-TM-MT and the~\textit{simplified}-$k$NN-MT models.}
  \label{tab:statistics}
\end{table}

\end{document}